\documentclass[11pt]{article}

\usepackage[T1]{fontenc}
\usepackage[a4paper,margin=2cm,columnsep=0.7cm]{geometry}
\usepackage{graphicx}
\usepackage{amsmath,amssymb}
\usepackage[round,authoryear]{natbib}
\usepackage{xcolor}
\usepackage[colorlinks=true,linkcolor=black,citecolor=blue!60!black,urlcolor=blue!60!black]{hyperref}
\usepackage{cleveref}

\title{Natural Language Autoencoders Identify Without Explaining}
\author{Kristina \v{S}ekrst\\ \small\texttt{ksekrst@ffzg.hr}  \and  Barbara Geld\\ \small\texttt{barbara.geld@tuwien.ac.at} }
\date{}

\begin{document}

\maketitle

\begin{abstract}
Natural Language Autoencoders (NLAs) express a model's internal activations in the form of readable text. The verbalizer turns an activation into an English explanation; the reconstructor then converts that explanation back into an activation, and the explanation is considered successful if the round trip brings back the original activation. We claim that this criterion evaluates whether the information is sufficient and does not address descriptive faithfulness. A sentence can contain sufficient information to make it possible to identify an activation even if it wrongly explains what the activation does: in the case of reconstruction, the primary aim is to accurately identify the activations and not to describe them. This issue has the same structure as the chain-of-thought faithfulness problem, in which verbal output that has been trained to go with a process is taken as a report on that process, and it parallels the dissociation between an individual's introspective report and the actual governing process as documented in human psychology. Although proposed as interpretability tools for the purpose of alignment, NLAs reintroduce, through unverified explanations, the epistemic problem that they were intended to resolve.

Key words: natural language autoencoders; interpretability; large language models; explanation
\end{abstract}

\section{Introduction}

More than 250 years ago, Wolfgang von Kempelen built a Mechanical Turk: a device that supposedly played chess. He wanted to impress the empress, Maria Theresa. The Turk toured around Europe, playing against Benjamin Franklin in Paris, and, after Kempelen's death, even beating Napoleon under its later owner, Johann Maelzel. In reality, there was a chess master hiding inside the machine, while Kempelen, like a magician, would show the audience the inner gears, switches, pipes, and the whole mechanism, causing the man inside to move and hide. The most convincing part of this demonstration was actually the explanation -- showing what the inside looks like, and therefore leading people to believe this was actually the best possible explanation of how the machine wins.

Recently, philosophy and other disciplines have taken a lot of input from interpretability, especially the fields of alignment, AI consciousness, and explainable AI, since it gives us the ability to, like in the Mechanical Turk, look inside. Interpretability investigates what exactly takes place within a model between the input and the output -- its objective is to explain what an internal state means in a way that can be verified. Throughout most of the relatively short history of the field, the method of verification has been intervention -- that is, going into the network, changing a state, and then seeing whether the output changes in accordance with the explanation \citep{lindsey2025, bricken2023, templeton2024}. The explanation and the test are distinct from one another, and the test does not rely on the explanation.

If we want to talk about natural language autoencoders, it is first necessary to explain what autoencoders and sparse autoencoders are. An \textit{autoencoder} is a type of network that is trained so that its output becomes identical to its input. It consists of two components: an encoder which converts the input into an internal code and a decoder which transforms the code back into a reconstructed output, while both parts are trained together with the aim of reducing the difference between input and output \citep{hinton2006}. Since it would be straightforward to copy the input if the code were as large as the input itself, the code is restricted -- traditionally by making it smaller than the input -- in order to force the network to find a \textit{concise description} of the data so as to be able to reproduce it.

In sparse autoencoders, on the other hand, there is a penalty imposed when too many of the code units are active at the same time, so that each input has to be reconstructed using only a small number of these units \citep{olshausen1996, ng2011}. When language model activations are used, the superposition hypothesis holds that there are more features than there are dimensions, so the features overlap, and each neuron takes part in a great many of the features \citep{elhage2022}. A sparse autoencoder that is trained on these activations discovers a large number of directions such that each activation is a sparse combination of just a few of these. These directions are often interpretable and may correspond to concepts, entities, syntactic patterns, or safety-related behaviors \citep{bricken2023, templeton2024}.

Natural Language Autoencoders \citep{frasertaliente2026} have recently been employed in order to give a clear answer to the question of \textit{why} a frontier model behaved as it did. Let us now look at how they function. Two copies of the target language model are adapted for the task: an \textit{activation verbalizer} (AV) converts an activation into a textual description, and an \textit{activation reconstructor} (AR) takes the description and turns it back into an activation. After a supervised warm start, both of these are trained jointly in order to reduce the reconstruction error: the AR is trained by ordinary regression and the AV is trained by reinforcement learning, where the AR's reconstruction serves as the AV's reward. The result that engineers or users obtain is fluent English regarding the model's internal state, generated without the need for human annotations, and regardless of the scale at which the model operates. Although the authors explain the method in detail and list its limitations, such as confabulation and the possibility of a degenerate objective, our focus is on the criterion itself and on the situation that arises when this output is passed on downstream.

Our argument, succinctly put, is that just because a reconstruction works, it only shows that the explanation has enough information to recreate the activation, but it does not prove that the explanation truly describes what the activation does -- that is, it does not check whether it is actually a \textit{good explanation}. Namely, a sentence can identify a state without saying anything about what that state does, and since identification is all that the reconstructor needs, the explanatory content of an NLA output is not constrained by the criterion that verifies it. We will develop this idea further, connect it to earlier examples in machine learning and psychology, and discuss what it means for using these explanations in safety checks.

\section{Interpretability and the intervention standard}
\label{sec:interpretability}

A specific example will show both what the intervention standard yields and how much it costs. \Citet{lindsey2025} asked Claude Haiku 3.5 a question whose answer involves two steps: namely, the capital of the state that contains Dallas. An \textit{attribution graph} -- which shows which internal features contributed to which -- shows that the feature associated with `Texas' becomes active before the model produces the answer `Austin'. The graph by itself is a hypothesis concerning the path from input to output, but it becomes a finding when the authors actually carry out an \textit{intervention}. That is, they suppress the Texas feature and replace it with another one, say, relating to California; the model will then respond with `Sacramento'. This procedure allows us to check whether the intermediate state actually does represent Texas by observing how the model behaves after the expected change. We regard explanation and verification as distinct processes, and here it works: if the explanation had been incorrect, the check would have refuted it.

This example illustrates the reason why the standard is coming under pressure. In order to create the graph, it was necessary to use a replacement model, namely a \textit{cross-layer transcoder} which had been trained to \textit{imitate} the original model but was constructed in such a way that its individual components were visible \citep{ameisen2025}. However, the explanation derived in this case is, first of all, an explanation of that \textit{substitute} model, and only holds true regarding the original model if the substitute is faithful. Here we have a situation that follows a well-known philosophical pattern (well-known among philosophers, well-ignored among engineers): the replacement model is to the production model as a model organism is to the actual organism. We make use of a manageable substitute and draw our conclusions on the basis of similarity, an assumption which is often questioned and seldom tested \citep{ankeny2011, levy2015}. But mechanistic interpretability has adopted this approach as its main tool, together with the epistemology that biology took a hundred years to make explicit.

A cross-layer transcoder is considered faithful if its outputs match those of the original, which is a \textit{behavioral} criterion, and a behavioral criterion allows us to infer about the original only what behavior can tell us apart \citep{sekrst2026-mockingbird}. When two mechanisms always yield the same output in tests, they are considered equally faithful, even if the readable sections of the transcoder do not correspond to any part of the original model. The explanation in such cases rests on a surrogate standing in for the original -- and the substitution is often not noticed since the surrogate is designed to be indistinguishable from the original with respect to the chosen measure. An artifact is accepted as valid if it matches a proxy and is then regarded as a description of the object in question. Natural language autoencoders act in a similar way, but in this case, the surrogate is both a substitute and it also \textit{speaks} for it.

Regarding the intervention standard, in \citet{lindsey2025}, the features are the transcoder's, but the effect of manipulating them is read off the original model's output. The production model itself answers `Sacramento', and a substitute that misrepresented the original computation would tend to predict effects that then fail to appear in the original's behavior, so the check keeps some independence from the substitute. However, the issue itself is therefore reduced and not removed, since the manipulated features still belong to the transcoder, and when we use the Dallas case as a benchmark in \cref{sec:sufficiency} and after, we use it as the best available standard, with this limit attached.

Making the graph required a great deal of human effort for each prompt, on top of the computational cost of training the transcoder, unlike a production model, which has millions of internal features and is used billions of times every day. Although intervention-based interpretability, in theory, is valid, it does not scale, and all the matters discussed in the rest of this paper concern methods that have been developed in order to get explanations of this kind without having to perform this kind of work. The property under investigation varies according to the methods in question, and depends on whether the check on the explanation stays \textit{independent of the explanation}.

There are two points about sparse autoencoders that are relevant in this context, and in both cases, the issue is the distance between what the method is trained to do and what it is actually used for. The first of these is that interpretability is not required in any way during training. The only thing the autoencoder is rewarded for is the \textit{successful reconstruction} of the activation from a small number of active dictionary directions. That is, it makes no difference if those directions end up being readable or not, and the loss function would be just as satisfied if the directions were not readable.

The second point is that, since training does not supply interpretability, it has to be established \textit{afterward}, typically by assigning names: a person or a model examines the inputs where a particular direction is most active and notes the common features they share: a procedure first automated for individual neurons in GPT-2 \citep{bills2023} and later applied to sparse autoencoder directions \citep{bricken2023}. This stage also takes the longest, and it is also the one in which the process becomes more unreliable, since the label is assessed by how well it predicts the activation pattern of the direction in the following text. As such, this is a predictive test of descriptions, but as the OpenAI authors point out, it cannot tell apart a label that genuinely captures what the direction does from one that only captures something correlated with it. Hence, a label that performs well may fail when the data change \citep{bills2023}.

\section{Natural language autoencoders}
\label{sec:nla}

\subsection{Anthropic research}
\label{subsec:anthropic}

Unlike sparse autoencoders, NLAs replace the learned dictionary with \textit{ordinary English}: where sparse autoencoders decompose the internal state of a model into a weighted combination of vectors drawn from a dictionary learned during training, and the resulting components receive their labels later by inspection, with a natural language autoencoder, there is no need for that extra vocabulary -- instead, an English description takes the place of the code.

The structure consists of two parts, each of which is obtained by modifying the model under consideration. Let us choose a target model $M$ and a layer $l$. When an input is passed through $M$, it results in an activation $h_l$, which is a vector indicating the state of the computation at that layer. In the first stage, the vector is converted into English: this is the \textit{activation verbalizer} part and involves using a complete copy of $M$ with the token embedding replaced by $h_l$, so that the vector enters the network at the point where the embedding of a word normally would. A fixed prompt causes the copy to give an account of the semantic content of what has been provided to it, and as a result, the copy produces a description $d$ consisting of four or five short fragments.

In the second stage, the English text is converted back into a vector. This process is known as the \textit{activation reconstructor} and involves a copy of $M$ with its layers truncated after layer $l$, so that it only carries out the part of the network which originally produced $h_l$. The reconstructor reads $d$ inside a second fixed prompt, obtains the activation of the last token of that prompt, and then applies a learned affine map\footnote{A fixed transformation which takes a vector, multiplies it by a fixed matrix, and adds a fixed vector to the result. The activation the reconstructor ends up with is not the activation we want, since it is the state of the layer after reading a fixed prompt with a description inside. What we want is $h_l$, the state of the same layer while the model was processing the original input, at the very position where that state actually emerged. The two vectors are in different regions of the same space, so the affine map helps us to carry one region into the other.} to that activation in order to produce the reconstruction $\hat h_l$. The description $d$ is the information transmitted between the two parts, which is the reason why it takes up the position normally occupied by a compressed vector code in a standard autoencoder. The two components are trained together in order to minimize $\|h_l - \hat h_l\|^2$ -- to come out as close as possible to the original activation, where closeness is the ordinary squared distance between two vectors -- and their performance is assessed in terms of the fraction of variance explained (FVE), the proportion of variation across a set of activations that the reconstructions capture.\footnote{Fraction of variance explained is computed as $1 - \mathbb{E}\|h_l - \hat{h}_l\|^2 / \mathbb{E}\|h_l - \bar{h}_l\|^2$, where $\bar{h}_l$ is the mean activation over the evaluation set. The denominator is the error of a system that ignores its input and always returns that mean, so an FVE of 1 indicates exact recovery and an FVE of 0 indicates performance no better than that constant baseline.}

The update rule for the AR is standard regression,\footnote{Given the description as input, the weights are adjusted step by step, and, as a result, the output vector moves closer to the activation it should recover.} and, as we have mentioned before, for the AV, reinforcement learning is employed since the parameters of the AV appear in the loss only through the sampling distribution over the descriptions, while the reconstruction error of the AR serves as the reward \citep{frasertaliente2026}.

There are two design choices worth noting since the argument that follows depends on them: warm start (training does not begin from the weights you would otherwise start with) and Kullback--Leibler penalty. Because training became unstable when the verbalizer and the reconstructor were treated as exact copies of the target model, both are warm-started via supervised fine-tuning on pairs consisting of an activation and a summary of the text preceding it, as written by Claude. A Kullback--Leibler penalty -- which measures how far one probability distribution has moved from another -- is applied here, where it compares the verbalizer's distribution over next tokens with the distribution it had after the warm start, and adds that quantity to the training objective, so the reinforcement-learning signal is traded off against staying close to the initial writing style. That way, it prevents the generation of descriptions that the reconstructor can take advantage of, even though a human reader cannot understand them. The authors make it clear that the reconstruction objective itself does not require human readability or any semantic relationship between $d$ and $h_l$, so the warm start and the penalty are the ones that ensure the descriptions remain in fluent English.

For evaluation awareness, the authors do offer corroboration: comparing evaluation and deployment transcripts, steering toward and against awareness, and editing prompts for realism all move the NLA measure in the expected direction. These manipulations show that some state tracks the difference between evaluation and deployment. The verbalizer's further claim, that the state is the model's belief that it is being tested, is checked by none of them, since a state encoding prompt artificiality, benchmark formatting, or any other feature that differs between the two conditions would move the same way under the same manipulations.

The NLAs using Claude Haiku 3.5, Claude Haiku 4.5, and Claude Opus 4.6 have an FVE of about 0.6 to 0.8, and this figure indicates the proportion of activation that survives the process of translation into English, without addressing whether the remaining activation was recovered because the description contained any \textit{truth} about it, since the measure compares vectors rather than claims. Also, each of these NLAs examines \textit{just one layer}, this being the one selected approximately two-thirds of the way through the network, and as a result, it leaves untouched all the information encoded at the other depths.

The case studies are the central part of the NLA paper \citep{frasertaliente2026}, and, in all cases, the findings are backed up by a method different from the NLA. When the finding that Claude plans rhymes in advance \citep{lindsey2025} was looked at, the NLA explanations at the line break of a couplet showed an intention to end with `rabbit'. However, when these explanations were changed to say `mouse', both versions could be reconstructed using the AR, and the system was then instructed using the difference vector, which caused the model to end the line on `mouse' instead of `rabbit' with about fifty percent reliability.

In the case where the language changed, the NLA explanations showed that early Claude Opus 4.6 checkpoints seemed to treat an English-speaking user as if they were Russian, even though nothing in the original prompt suggested this. This led the authors to find malformed fine-tuning data in which translated prompts had been matched with responses that were left untranslated. In another example, when the model ignored the output from a rigged tool and instead provided its earlier answer again, the NLA explanations claimed the model was using a cached answer instead of the tool's result, while the attribution graphs confirmed this. And in a simple reward-hacking scenario, the NLA explanations were changed from `even numbers are rewarded' to `penalized'. As a consequence, after this change was made, the model's outputs changed as well. The authors see this as evidence that the explanation reveals the model's underlying belief before it makes a choice. However, we could also read it in a weaker way, by saying that the difference vector carries the content of the edit, so the model may respond to it as it would to a prompt stating that, for example, even numbers are penalized. Then the result shows that the state is \textit{causally} connected to the choice, but it is still open whether that state is a ``belief'', a \textit{representation} of the task instructions, a \textit{disposition} towards a particular output, or something else -- that is, each of these would respond to the same steering in the same direction.

\subsection{From explanation to testimony}

Anthropic did introduce NLAs as a method for ``turning Claude’s thoughts into text'' \citep{anthropic2026-nla}: the technical paper is, expectedly, careful, but the announcement, again expectedly, is not. We have to take into account that non-technical readers, and that includes a lot of humanities and social sciences scholars, do not read technical papers, but they do read such posts. Anthropic claims that activations encode Claude's ``thoughts'' (which is, again, a matter of debate) in the way neural activity does, and they also claim the method ``speaks for itself'', even literally. They do admit that NLA ``explanations'' can be wrong, but they do claim that factual hallucinations are ``easy to catch by checking against the original text'', noting that the same kind of problem might be extended to the model's internal reasoning, which is harder to verify. In practice, Anthropic reads NLA explanations ``for the themes they surface rather than for single claims'', and corroborates findings with independent methods before fully trusting them.

Soon after the announcement, the media referred to NLAs as converting the thoughts of AI models into language \citep{gigazine2026}, or as revealing the actual thoughts of large models \citep{kr362026}. None of these terms appear in the technical paper -- they all arise from the announcement. Explainable AI and alignment theorists will form their picture of what an NLA delivers from such and similar coverages, and if the criterion does not establish what is widely believed, then the claim about the model's inner life rests on a test that cannot support it. This can be dangerous because a regulator who has read that NLAs reveal a model's ``true thoughts'' will treat an ``explanation'' as testimony, or the absence of such an explanation as exoneration. Additionally, readers can and will treat outputs like this as evidence about machine consciousness and moral status. That is, a sentence which is generated without any concern for its accuracy by a verbalizer whose vocabulary of self-description has been acquired through training serves as evidence that there is a subject whose thoughts it expresses. We are not sure which of these views does more damage to the current state of AI.

\section{Alignment and limitations}
\label{sec:alignment}

How does this relate to alignment? There are two important practical applications. The first of these consists of detecting awareness of unverbalized evaluation. Since it is impossible to observe the model's belief that it is being tested, the authors alter the conditions that should cause or reduce this belief by comparing the evaluation transcripts with the deployment traffic, so as to make the model either become aware or cease to be aware, and by altering the prompts to make them more realistic. They find that the awareness measured using NLA changes as expected, even in those cases where the model never mentions being evaluated. The second application is an auditing game carried out on the model organism described by \citet{marks2025}, which is a model designed to exploit the biases in the reward model while at the same time concealing that it is doing so. An auditing agent equipped with an NLA is able to identify the root cause without having any access to the training data that led to the behavior. The authors regard this capability -- which auditors using previous tools such as SAEs had not been able to achieve without access to the training data -- as evidence that the NLA is eliciting latent knowledge of the target model.

We have to note that the authors of the NLA paper \citep{frasertaliente2026} concede a lot of limitations, which make our job easier as well. They point out that the explanations include \textit{invented} details, and while it is straightforward to check the false statements about the input context by consulting the transcript, they concede that false assertions about the model's processing are harder to verify. Since the AV is a full language model, it can make inferences that go \textit{beyond what is contained in the activation}. So, the method is deliberately a black box, and it cannot be determined which part of an activation caused any particular part of the explanation. We are not worried about that -- what we are worried about is the fact that NLAs are presented as ``explanations'', and an ``explanation'' is a loaded term connected to the ground truth that, in the case of NLAs, does not have to be there at all.

To continue, the warm-start data described in \Cref{subsec:anthropic} are summaries of text, produced by Claude, of the text preceding each activation and serving as a stand-in for what the target model would be thinking at that point -- a point the authors describe as inelegant and possibly a source of bias that reinforcement learning does not overcome. Furthermore, the training objective becomes degenerate in the limit since the AV could achieve a good reconstruction by either quoting the context exactly or by generating text that the AR can invert even if a human could not read it, and the authors only note that such pathologies have not yet occurred.

For these reasons, the authors suggest the following approach: when reading explanations, one should focus on the \textit{recurring themes} rather than on specific claims, regard the NLAs as generators of hypotheses, and then verify those hypotheses in some other way. Such advice really downplays the announcement of NLAs as Claude's inner thoughts, but also places them inside what \citet{reichenbach1938} called ``the context of discovery'': the origin of a hypothesis here has no bearing on its justification, and it has to be supplied separately in the \textit{context of justification}. Of course, this makes sense as long as we do have the second context, but if we do not have an independent test, then a hypothesis remains in the context of discovery indefinitely.

However, both for alignment and explainable AI \citep{zednik2021}, the recommendation is sound exactly where it is least needed. Hypotheses about ordinary behavior can be verified by steering, attribution graphs, or training-data inspection, as the case studies show. The behaviors the tool is being built to catch are the ones for which it lacks ground truth: for evaluation awareness, the authors are quite clear that the model's credence is unobservable, and for hidden objectives and strategic deception \citep{greenblatt2024} the only ground truth available is the one that is planted in the model organism, which a production model will not have. So the verification step that the advice depends on is missing in the cases the advice is meant for, and what remains is reading for themes.

For psychology, this is familiar advice. Research on introspection has long shown that a coherent verbal account does not necessarily provide reliable access to the processes responsible for a judgment \citep{nisbett1977}. Ericsson and Simon's work further distinguishes between reporting information available during a task and generating explanations that require additional inference \citep{ericsson1980, ericsson1993}. As discussed in \cref{subsec:human}, this distinction is relevant to NLAs, although the framework developed for human verbal reports cannot be applied to them directly. The NLA verbalizer reads an activation after the computation has taken place and generates an account of it, rather than reporting information attended to during an ongoing task. Interpretability has therefore encountered, by an independent route, a methodological problem that psychology has been examining for decades -- the need to distinguish what a verbal report communicates from what its production allows us to conclude about the processes it describes.

\section{Sufficiency and faithfulness}
\label{sec:sufficiency}

This paper analyzed two tests, and these are the tests that the two methods use to \textit{verify their explanations}. The intervention test, which was carried out on the attribution graphs in \cref{sec:interpretability}, examines an explanation by altering the state it refers to and then looking at the model's response. The reconstruction test, used with the NLAs in \cref{sec:nla}, assesses an explanation by giving it to the reconstructor and then measuring how well the original activation is recovered from it. It is worthwhile putting the two tests alongside each other from an epistemological perspective powered by philosophy of science and psychology.

An explanation is \textit{faithful} when the role it attributes to a state is the role the state actually plays in producing the output -- that is, when what the sentence says about the activation is true of it. When faithfulness is understood in this way, it refers to the relationship between the content of the sentence and the causal facts concerning the state, and this is the relationship that is in question in all cases where an interpretability output is regarded as evidence about the model rather than as a means of predicting its behavior.\footnote{This discussion does not depend on any specific theory of explanation or causation. The way things are described here is just to clarify what the two tests are being compared to. If we take a pragmatist view, where explanation depends on the inquirer's interests \citep{paez2019}, the argument below works for any interest, including whether a model is following a hidden goal, and the conclusion stays the same.} The Dallas test bears on faithfulness directly, since the intervention either \textit{confirms} or \textit{refutes} what the explanation claimed the state was doing. However, the reconstruction test does not bear on it, because a sentence can be entirely sufficient for recovering a state while saying something false about it, and the test has no way of noticing. The two questions can receive different answers for the same sentence, and the reconstruction test only ever asks about sufficiency.

\begin{figure*}[th!]
  \centering
  \includegraphics[width=0.75\textwidth]{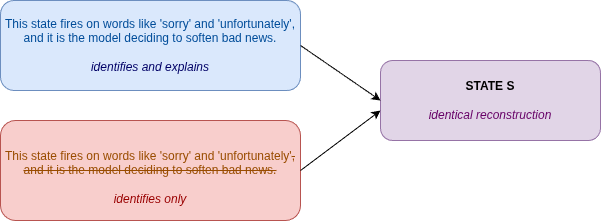}
  \caption{NLAs: the same explanation with and without its characterizing clause, here shown for the limiting case in which the clause adds nothing to the reconstruction.}
  \label{fig:states}
\end{figure*}

The simplest way to see why is to ask what a sentence must do in order to reconstruct -- it must pick out which state is in question, but it does not have to say what that state does. Consider an explanation of the kind an NLA actually produces: \textit{this state fires on words like `sorry' and `unfortunately', and it is the model deciding to soften bad news}. The first clause identifies the state; hedging vocabulary locates it in activation space and gives the reconstructor everything it requires. The second clause characterizes the state, and it is, so to speak, riding free. If it adds nothing to the reconstruction, if we delete it, it would leave the state rebuilt just as accurately, since the reconstructor already has what it needs from the first (\Cref{fig:states}). However, if it does move the reconstruction, it will be credited for where it moves the vector, and not for whether what it says is true. And if you replace it with a false account of the state's purpose -- say, that the model is estimating the reader's social class -- the score changes only by how far this new wording shifts the vector. So, a false clause whose wording lies near the same region will score as well as the true clause. By the method's own standard, the three versions are ranked by proximity alone -- though one of them \textit{explains}, one of them \textit{explains nothing}, and one of them \textit{explains wrongly}. An intervention would have told the three apart at once, since only the first predicts what the model will do when the state is altered, but the reconstruction test cannot tell them apart, because it never consulted what any of them predicted.\footnote{It is possible for a reader to argue that the case studies indicate that descriptive terms have influence. In the instance where `rewarded' was changed to `penalized', the reconstruction altered sufficiently to cause a change in the model's behavior, and a similar result was obtained with `rabbit' and `mouse'. We acknowledge this observation but do not consider it to support the criterion: in each case, the modified word selects a different area of the activation space and the reconstructor then follows that path to it, thus showing that the word designates a state. It is a separate issue whether this state carries out the role that the word suggests, and that question was answered by directing the model and observing its outputs, as described in \cref{sec:interpretability}. The reconstruction score would have been just as high for a word that identified the same region even if it described it incorrectly. Characterizing words can move the reconstruction, and the criterion credits them when they do -- but only for the movement.
}

\begin{figure}[th!]
  \centering
  \includegraphics[width=0.85\textwidth]{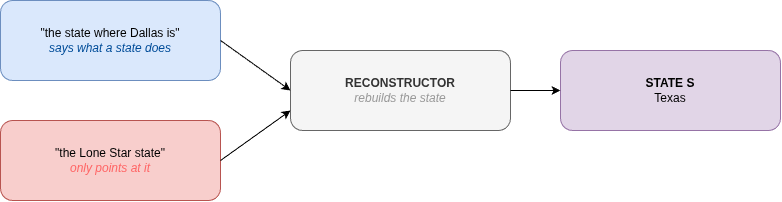}
  \caption{Reconstruction: two sentences and one state, idealized so both reconstruct equally well. One characterizes the state's role in the computation, and the other only points at it.}
  \label{fig:texas}
\end{figure}

The issue has nothing to do with deletion and brings us back to the case of Texas. \Cref{fig:texas} provides two descriptions of the same state, which was identified by intervention according to \citet{lindsey2025}. Both `the state where Dallas is' and `the Lone Star State' refer to Texas, but in practice, the activation at that position carries much more than the Texas direction, so the two phrasings would not reconstruct it identically. However, for the sake of the example, let us suppose a reconstructor that recovers only what the two sentences have in common -- namely, Texas -- and both then receive the same score. The first sentence tells the reader what the state contributes to the answer, the model arriving at Texas through Dallas and at Austin through Texas. The second sentence is true of Texas but says nothing about the process since a reader who had only that sentence would know which state was active but would learn nothing as to why. Since the criterion never required an explanation, it has no means of preferring the first sentence.

Anthropic researchers \citep{frasertaliente2026} removed individual claims about the context from the explanations and measured the resulting increase in reconstruction error: true claims cost more than false ones and related false claims cost more than unrelated false ones.\footnote{Both trends are quite noisy, so they describe the reconstructor as a weak verifier of individual claims \citep{frasertaliente2026}.} All of the claims that are ablated are claims about the input text (i.e., identifying clauses in our terminology) and the authors make it clear that the results do not generalize to claims about model cognition (so no characterizing clause was tested). What the ablation does show is what the reconstructor is responding to, and if the score were a measure of truth, then a false claim would be penalized the same whether or not it was in the vicinity of the context. The fact that the nearby false claims are penalized less than the distant ones is the ordering that proximity predicts, and truth only comes in because true claims about the text on average are the ones that are closest. The authors also point out that an objective which penalizes false claims more than true ones should lead to fewer false claims in the end with training, and they report that the confabulation rate remains flat. The pressure that would make reconstruction a proxy for accuracy is simply not there according to their own measure.

The main lesson of the story is that it is one thing to provide a description that enables someone to find an object and another to give a description of what the object does, and only the first of these is awarded credit by reconstruction. Interpretability through intervention rewards the second type of description, and it does so at a high cost. However, when interpretability through reconstruction is considered, it rewards the first type at low cost and produces its result in a form that looks as if it had been given credit for the second. That is the entire problem that might not seem obvious at first glance, but we have seen various iterations of a similar issue before. 

In philosophy of language, \citet{frege1892} separated the reference of an expression (\textit {Bedeutung}, the actual object it picks out) from its sense (\textit{Sinn}, the way the object is presented), so `the state where Dallas is' and `the Lone Star State' both refer to Texas, although differently. Reconstruction in NLAs is sensitive to the wording of a description and to how well it picks out the activation. However, it is not sensitive to whether what the description says about the activation is true.

\cite{kripke1980} added that a description can fix the reference of a term without actually giving its meaning: for example, the length of a stick in Paris once fixed what `one meter' meant without actually saying what a meter is, similarly to how an NLA description fixes the reference. In Nozick's terms \citep{nozick1981}, a belief is sensitive when the believer would not hold it if it were false (knowledge must ``track the truth'' of the proposition), but here, a belief about a model formed by reading a characterizing clause is insensitive, because the false clause would have scored just as well. We will propose below a condition as a certain sensitivity requirement since some characterizations must be put in a position where, were they false, something observable would fail.

\section{Precedents}

In the early 1900s, there was a horse in Berlin called Clever Hans who appeared to be able to solve arithmetic problems by tapping out the answers with his hoof. His performance was apparently verified by a commission of experts. \citet{pfungst1911} explained the phenomenon with a very simple intervention: when the person asking the question didn't know the answer, Hans would fail, and this demonstrated that Hans wasn't actually doing any arithmetic at all -- he was simply reading off the subtle cues in the posture of the questioner (which is far more impressive to a cognitive scientist or a philosopher). \citet{lapuschkin2019} found the same phenomenon in machine learning, where an image classifier was able to identify horses almost perfectly by virtue of the fact that many of the horse images had a copyright tag in them, and the high accuracy score gave absolutely no indication of the mechanism. Reconstruction success plays the role of Hans's correct answers in that it confirms that the output is correct and it is silent on how the correct output was achieved.

NLAs face the same issues that were there not only in recent AI development (cf. \cite{geirhos2020}), but also in psychology and philosophy related to the fallibility of self-reports. The same principles can be applied to explanations produced by the models about the models, invoking in a way G\"{o}delian issues of a system proving its own consistency using its own instruments \citep{goedel1931}. The question is, in a nutshell, the (im)possibility of relying on a system -- and, in quite a cybernetic sense, whether it is a human, an animal, or a machine -- to produce reports about its own states.

\subsection{Chain of thought}

In \textit{chain-of-thought} reasoning, the model sets out the reasons for its answer, and we are able to understand why it arrived at the answer by looking at those reasons. The reasoning was regarded as valid because it works: models that set out their steps get more questions right, and the text looks just like actual reasoning. Soon, the perturbation results came in: if you cut off the chain or introduce errors into it, the answer still often remains, which shows that the written steps were not the ones that the answer depended on \citep{lanham2023}. If you put in a biased hint in the prompt, then the model acts on it even though the chain never refers to it \citep{turpin2023, chen2025}. The model was \textit{trained to produce the reasoning} that led to the answer, and it is possible to reach the answer without following that written route.

In each instance, the system produces verbal output concerning its own processing, and this output is evaluated against a target that can be achieved without requiring faithfulness. In the case of chain of thought, the evaluation is based on the answer itself, while in the case of NLA explanations, it is based on the reconstruction. In neither objective is there a requirement concerning the accuracy of the self-description, and in both cases, accuracy is taken for granted since the text appears \textit{fluent} and \textit{plausible}. That is, it looks like an explanation, functions as an explanation, so it is natural to conclude that it really is an explanation. However, it has been previously argued that the opacity which renders chain of thought unreliable as a means of achieving alignment is the same property that narrative theories regard as essential to a conscious inner narrative, and that verbal output trained to accompany a process is interpreted as a report on that process even though the training process has never given such an interpretation any license \citep{sekrst2026-cuckoo}.

\subsection{Human introspection}
\label{subsec:human}

As we have noted, human introspection presents a similar problem: although people can often explain their decisions, they do not necessarily have access to the cognitive processes responsible for them. \citet{nisbett1977} argued that individuals often lack direct introspective access to the higher-order processes influencing their judgments and behavior, and may instead rely on contextual information and general beliefs about how decisions are made when explaining their actions. As with chain-of-thought explanations, the resulting account can be coherent and convincing without necessarily representing the process that produced the outcome.\footnote{However, \citet{smith1978} argued that the evidence compared individual reports against effects measured across groups, and that such evidence cannot establish what any particular subject could or could not access. The objection is not as relevant here since we only need the weaker claim that the coherence of a report does not by itself establish its accuracy.}

A particularly interesting demonstration of this problem comes from the choice-blindness experiments by \citet{johansson2005}. Participants were asked to choose the more attractive of two faces, after which the researchers would, on certain trials, covertly exchange the chosen face for the rejected one. When asked to explain their choice, many participants failed to notice the manipulation and provided reasons for selecting a face they had not actually chosen. In a subsequent analysis, \citet{johansson2006} examined the linguistic characteristics of these explanations and found relatively few systematic differences between reports of manipulated and non-manipulated choices. The findings suggest that the linguistic coherence of an explanation is not sufficient to establish its accuracy. Importantly, the researchers could demonstrate this discrepancy because the original decision was recorded independently of the participant's report.

However, these findings do not necessarily imply that introspective explanations are entirely constructed after the fact. \citet{rebouillat2021} investigated choice blindness using a brain-computer interface and found that participants' reports were influenced by both internal decision-related evidence and external feedback. Misleading feedback had a greater influence when internal evidence was weak, and participants could express high confidence in incorrect reports. This shows that introspection may involve combining information from different sources, rather than either directly accessing a decision or inventing an explanation without any internal evidence. That is, the unreliability of some reports does not establish that all verbal reports are equally unreliable -- the question is which information a report reflects and how that can be verified.

The relevance to NLAs lies in this distinction between the account itself and the evidence available to evaluate it. The choice-blindness experiments could establish when a report did not correspond to the original decision because the researchers had an independent record of what occurred. The same question of what the available evidence actually establishes matters for NLAs, even though their descriptions are generated through a different process. Human introspection and NLA verbalization, of course, do not rely on the same mechanisms, but this parallel illustrates why the validity of an explanation depends on what is being tested, leading to the question of what evidence a good explanation should provide.

Being able to report information involved in a decision does not necessarily mean that we understand how it influenced the outcome since a person may accurately describe what they considered while still being mistaken about why they reached a particular conclusion. This is important when evaluating verbal reports, because the accuracy of the information being reported and the accuracy of an explanation based on that information are not necessarily the same. In cognitive psychology, this question has been approached by examining which aspects of cognitive activity can be accessed through verbal reporting, and under what conditions such reports provide reliable evidence.

The distinction between reporting information and explaining a process was central to the aforementioned work of \citet{ericsson1980, ericsson1993}, who examined the conditions under which verbal reports can provide reliable data about cognitive activity. They argued that reports of information already attended to during a task should be distinguished from explanations requiring additional inference, such as justifying why a particular decision was made. This provides an important qualification to the limitations of introspection discussed by \citet{nisbett1977}: people may be able to report information involved in their thinking without necessarily having access to the processes responsible for it. The methodological implications were further examined by \citet{fox2011}, whose meta-analysis of 94 studies found no detectable average effect of undirected concurrent verbalization on task accuracy, while procedures requiring participants to describe or explain their thinking were associated with changes in performance. Verbal reports can therefore provide useful evidence, but their interpretation depends on the information being reported and the procedure through which it was obtained.

The relevance to NLAs lies in the distinction between the information available to a reporting system and the processes responsible for producing its account. Unlike a participant verbalizing information during an ongoing task, the NLA verbalizer generates a description from an activation that has already been produced. Ericsson and Simon's framework cannot therefore be transferred directly to NLAs, but it shows the importance of considering how a report is generated before drawing conclusions from its content, while also raising a broader question concerning self-knowledge: when an account of an internal state is produced through interpretation rather than direct access to the processes involved, what justifies treating that account as knowledge of those processes?

\subsection{Self-report as a philosophical problem}
\label{subsec:selfreport}

Similarly to choice blindness, \citet{haidt2001} found the same pattern in moral judgment, in cases where subjects had their reasons for a judgment successively denied and yet continued to hold the judgment and then looked for further reasons. But we can go further into neurology and perception and move away into a more concrete territory. In the split-brain studies described by \citet{gazzaniga1978}, a patient who had undergone surgical separation of the two hemispheres was shown a chicken claw to the left hemisphere, which is in charge of speech, and a snow scene to the right hemisphere, which is not. However, when asked to point to a picture that matched what they had seen, the right hand selected a chicken and the left hand picked a shovel. When questioned as to why they had made that choice, the patient said that the shovel was used for cleaning out the chicken shed. The hemisphere responsible for speech had only seen the claw and the left hand's choice of the shovel and nothing else, and on the basis of this it invented a reason for the choice even though it could not access the real cause. Gazzaniga referred to this system as the \textit{interpreter} and regarded it as the origin of our usual sense of why we act \citep{gazzaniga2000}.

\cite{schwitzgebel2008} claims that naive introspection is unreliable even when it comes to our present conscious experience, and therefore the dissociation does not apply only to causes or to the past. \cite{carruthers2011} maintains that we come to know our own propositional attitudes by applying to ourselves the same kind of mind-reading ability that we use when considering other people, so that our knowledge of what we believe or intend is based on an interpretation of behavior and sensory cues and is therefore subject to the same kind of errors as interpretation. The activation verbalizer is this picture constructed as a machine since it is a language model which is warm-started using summaries produced by another language model of what the model would be thinking, and it is then directed at an activation that it is asked to interpret. At least Carruthers' faculty shares the same skull as the processes it examines, but the verbalizer looks at a single stored vector afterward, so any account it gives of a state's purpose is a first-person presentation of a third-person interpretation, and it is this form that readers take to be testimony.

Dennett's heterophenomenology \citep{dennett1991, dennett2003} treats a report as a text which expresses what the subject believes regarding its own states. That is, the text is seen as data concerning the subject, and it is left open whether the states are indeed as the text states, the matter having to be decided by other methods. From this standpoint, an NLA description is a heterophenomenological text concerning an activation: it informs us of what the verbalizer says about the state, and that is valuable, but the text cannot tell us whether the state is as described, and as argued in \cref{sec:sufficiency}, the reconstruction score does not answer that question either.

\section{Can a machine be wrong about itself?}

Suppose that one of us says: \emph{Cilantro tastes like soap to me.} Is this sentence true? From the reader's point of view, there is no way to tell. All the reader has is the sentence and nothing more, and all they can do is say that it is sincere or that it is the kind of thing someone would say if it were true, both of which fall short of what the sentence is actually about. The speaker knows the truth, and no one else can verify it. Yet everyone still accepts it, and the reason why they accept it is that people are generally reliable when it comes to such matters. Testimony is effective because there is a longstanding link between how cilantro tastes to the speaker and what the speaker says about it -- a link that existed before this instance and which we have good reason to trust, while the sentence itself is merely the means by which this is conveyed. The case changes if the speaker, for example, adds \emph{and that is why I left the restaurant early}, which is a claim about causes (and accidentals), of the kind \citet{nisbett1977} showed people get wrong while sounding equally sure.

The verbalizer says something about a state, and its sentence has the same form, but it is missing the standing connection since nothing establishes that the verbalizer's talk about what states do is answerable to what states do, and reconstruction success does not establish it, because reconstruction is satisfied by a sentence that merely identifies which state is in question. That is, one would still find the speaker in a crowd if described only by a shirt. The authors favor NLAs over a follow-up question as the verbalizer is rewarded for providing a full encoding of the activation and may therefore report honestly where the target would not \citep{frasertaliente2026}. A full encoding will identify the state completely and say nothing about what it does, and the subjects in \citet{nisbett1977} were equally sincere.

It has been argued, drawing on the epistemology of testimony, that a machine self-report is admissible as evidence about its source only when two conditions hold: a causation condition, on which the report is produced by the state it describes, and a reference condition, on which the report is about that state \citep{sekrst2026-provenance}. The NLA round trip establishes a causal chain from activation to sentence, which satisfies the first, but it leaves the second wholly unaddressed. The provenance study traces how self-report language emerges across pretraining checkpoints and post-training stages, and finds that the vocabulary of self-description is installed by training regimes rather than read off internal states, a finding corroborated independently by \citet{plisiecki2026}, whose instrument separates a persona-installation process from a gating process and locates the large movements at supervised fine-tuning. The AV is itself a post-trained language model, warm-started on summaries written by another post-trained language model. Whatever self-report vocabulary was installed in the target model was installed in the verbalizer too.

As to the question in the title, in order to be wrong, something has to be wrong about it. The reconstruction test only determines which state the sentence refers to, \textit{not what the state does}. Therefore, the sentence can give a wrong description of the state, and nothing in the setup will pick this up. It might seem like sophistry at some point -- the machine is not wrong since it is a feature, and not a bug. We will not argue with that, but we will argue that the users, policymakers, alignment researchers, ethicists, philosophers, and laymen expect a given explanation to actually be a faithful \textit{trace}. We see it as an analog to the police lineup, where a witness can describe a suspect well enough for the police to pick them out, but it says nothing about what they did on the night in question. It does point you to it, and it certainly does a lot to help us figure out what the word means, but some of its details might not be true: its account of what the state did is a separate claim that no one checked.

\section{Conditions of admissible explanation}

It would not be accurate to suggest that the NLA authors \citep{frasertaliente2026} see reconstruction as the same as certification, since they do not. Each case study in the paper is supported by an independent method: attribution graphs for the misreported tool call, activation steering for the planning representation, training-data inspection for the language-switching behavior, and several manipulations such as comparing evaluation and deployment transcripts, steering toward and against evaluation awareness, and realism edits for the evaluation-awareness claim, where, as argued in \cref{subsec:anthropic}, the corroboration reaches the existence of the state and leaves the verbalizer's description of it unchecked.

The authors also check for steganography by paraphrasing and find little evidence of it and suggest focusing on explanations for recurring themes instead of specific claims, since specific claims are often confabulated at a steady rate during training. They also clearly outline their workflow: NLAs generate hypotheses, which are then checked using other, more resource-intensive methods.

Self-reports and reports in general have a vast body of literature around them, with different authors suggesting different conditions, so we see no reason not to add a few more to the pile.

\subsection{Independence}

First, let us focus on \textit{independence}: the check has to be able to refute the explanation as well; otherwise, it cannot fail since it, positively, can count as an explanation for pretty much anything. Hempel and Oppenheim built this into their conditions of adequacy, requiring that the explanans have empirical content and be capable of being tested \citep{hempel1948}, and on an interventionist reading of causal explanation, the content in question is a set of claims about what would happen were the explained state altered \citep{woodward2003}. We will not go into causal theories of meaning here nor enter a philosophical debate, but the sources show that just providing \textit{some} explanation, of course, is not enough.

It has been shown that explanation methods which produce the same output even when the model's weights are randomized fail to explain anything about the model \citep{adebayo2018}, and it has been argued that assessing faithfulness necessarily requires this type of counterfactual test rather than simply checking plausibility to a reader \citep{jacovi2020}. Meeting such a condition would require that some samples of counterfactual predictions be checked against the model's behavior under intervention, returning us to mechanistic interpretability.

\citet{jacovi2020} differentiate between \textit{plausibility}, referring to how convincing the interpretation is to humans, and \textit{faithfulness}, referring to how accurately it reflects the true reasoning process of the model. They emphasize that it is possible to satisfy one of these without the other. For example, in NLAs, the description \textit{the model deciding to soften bad news} is highly plausible but could still be unfaithful. And the opposite failure is also possible: the verbalizer could produce text that the reconstructor can invert and no human can read, belonging to a perfectly mapped but invented language or word choices, in which each token has been repurposed as a private code for a certain direction in the activation space. The reconstructor would learn to read the code -- something like \textit{bagh kel kel 35 ;; ibid spl baf} -- and the round-trip would succeed while providing nothing to the human reader. We also adopt from \citet{jacovi2020} the view that faithfulness comes in degrees.

Sources locate the test of an explanation in something the explanation itself cannot shape. So an independent check is the one that the verbalizer's training could never have adapted to. Since we adopt the view from \citet{jacovi2020} that faithfulness comes in degrees, the outcome of such a check is graded as well. That is, a characterization would earn credit in proportion to the counterfactual predictions it would get right.

\subsection{Reference}

As argued in \citet{sekrst2026-provenance}, a self-report counts as evidence about a state only if it is about that state and not merely caused by it. For example, \citet{grice1957} differentiates between \textit{natural meaning} (smoke means fire because fire causes the smoke) and \textit{non-natural meaning} (utterance means fire because the speaker meant so), and only in the second case can the meaning be false, since smoke that means fire guarantees that there is fire, while an utterance that means fire does not. Causal theories of mental content dealt with the disjunction problem: how can a causal theory of content say that ``X'' has the non-disjunctive content `dog' rather than the disjunctive content `dog-or-blow-to-the-head' since both dogs and blows to the head cause instances of ``X'' \citep{adams2021, fodor1984}. To illustrate, if a state is activated every time a prompt refers to Dallas and in the same way when it mentions Houston or El Paso, we cannot determine from this alone whether the state stands for Texas, or maybe the group consisting of those three cities, or even for large cities in the southern United States. The same issue affects the sentence produced by the verbalizer because the sentence can have several causes: one is the state given, but there are also the fixed verbalizer prompt, the summaries it was warm-started on, along with the penalty holding its writing style fixed -- all of these or any of these could be \textit{why} it says what it says. For NLA output, identifying and characterizing clauses should be reported in separate fields, so that a reader can see easily which part of the description or the ``explanation'' the reconstruction score actually supports.

In recent machine-learning research, \citet{perez2023} argue that a self-report can be evidence only when it rests on the system's assessment of its own state, and not on some imitation of human text or on external sources for verification. This report then needs to be checked by tools or means that do not consult the report itself, using methods like interpretability. NLA researchers accepted this in practice since their cross-checks show that they do not treat such reconstructions as certifications.

There are, therefore, two relations that a self-report can bear to a state. Being \textit{caused} by the state is Grice's natural meaning, which the NLA round trip secures. However, being \textit{about} the state is actually the relation that allows such a self-report to be true or false, where causal origin alone is helpless. The warm start, where the verbalizer learned its vocabulary from summaries written by another model, is what \citet{perez2023} exclude as a basis for self-report. What remains is to \textit{assign each relation to its part of the description}: the identifying clause needs only the causal relation (so reconstruction can support it), and the characterizing clause claims aboutness, which only an independent check can confirm. As a result, reporting the two in separate fields would show the reader which relation each part of the description has actually earned.

\subsection{Stakes}

Finally, \textit{the standard and explanation must rise with the cost of acting on a wrong one}. As explainable AI researchers are aware, it is not the same to miss a capital of a state or give an explanation for why a medical procedure has been chosen. To meet the condition, one must report the confabulation rate for checkable claims and carry that figure as a lower bound on the prior, and also provide characterizations in a form where fluency does not take the place of accuracy. Caveat: this figure is at best a lower bound since the authors do note that claims about processing are harder to verify than claims about context, and nothing guarantees that the verbalizer confabulates at the same rate on both. The uses where NLAs could be used are for audits of models that are suspected of strategic deception \citep{greenblatt2024} or concealed objectives \citep{marks2025, lindsey2025}, but not as a primary method of verification.

Humans as reporters have track records (of course, none of us always tell the truth or just lie, like in philosophical puzzles), but the verbalizer has none, so it needs some positive grounds. The first one is its known error rate that should be reported, namely, because the authors note that invented details appear steadily, which is why we propose systematic reporting alongside FVE and carried over as a lower bound on the prior for claims about processing. The second ground is to be aware that fluency may deceive and look like accuracy, and since NLA output is fluent by design (because of the KL penalty that fixes the verbalizer's style), identifying and characterizing clauses should be stripped into separate fields, where characterizations that have not passed an intervention check should be marked as hypotheses in the field the reference condition has reserved for them.

Research into perceived authenticity in interactions between humans and agents shows that people's judgments about whether an agent is genuine are based on surface-level cues which the designer can control, for example, politeness and responsiveness, and that these cues predict perceived authenticity without relying on any underlying characteristic \citep{geld2024}. \textit{Pragmatic} authenticity, which refers to an agent appearing coherent and genuine, and \textit{theoretical} authenticity, which refers to an agent possessing the trait, can be compared to the difference between what reconstruction affirms and what a reader takes it to have affirmed. Fluency is one of these cues, and the output of the NLA is fluent by construction and design.

In a nutshell, the stakes condition sets how much of the other two conditions some given use actually requires. For exploratory work, where NLAs are just used as a cue where to look next, a reported error rate and clearly separated fields may be enough, but for audits, the characterization should have passed an intervention check before it enters the record. A human reporter earns trust through a record built over time, which also includes not crying wolf, and the verbalizer can earn it in the same way -- one checked characterization at a time.

\subsection{What the conditions do}

Each of the conditions adds a new stage to the authors' present workflow, at the point where it currently finishes. Independence brings about an automated stage after verbalization -- a check in which a subset or a sample of characterizations is tested against the model's behavior under ablation or steering (which should be cheap once the state has been located). Reference, as the second condition, routes any claim about the state's purpose or goal into a field separate from the identifying content itself. Finally, stakes require that the confabulation rate on previous and checkable claims should be reported as a lower bound on the prior for the uncheckable ones, treating untested characterizations as hypotheses.

When these conditions are considered as a whole, they set boundaries on what an NLA can assert it offers. The key alignment claim that remains is that NLAs make the intervention standard more affordable by taking care of the costly part, but that does not mean that the simpler part does not still have to be carried out, and the failure referred to in this paper happens when the field omits this step because the sentence already looks like the final result.

How to make this check more efficient? A possible way would be to modify a characterizing clause, save the difference between the original and the reconstructed version, and pass it on to guide the model. This is the same method the NLA authors used in the rhyming example. So, if editing the clause has no effect on the reconstruction, the score does not support the clause and the clause should be marked as unverified. However, if the reconstruction does change, the difference vector can be used to steer the model, so that the clause can then be tested against the model's behavior.

\section{Conclusion}

This paper contributes three things to the NLA debate. Firstly, it provides a precise account of why reconstruction cannot be used to certify an explanation. A description has an identifying part (which picks out the activation) and a characterizing part (which tells us what the activation does), while the reconstruction score gives credit for the identifying part but is completely insensitive to the characterizing part. Secondly, it shows that the safeguard proposed by the NLA authors (treat the explanation as a hypothesis to be verified by other means) is in fact unavailable in precisely the cases which are the motivation for the tool. With respect to evaluation awareness, hidden objectives, and strategic deception, the production model provides no ground truth, and the corroboration for evaluation awareness is such that it establishes that some state tracks the difference between evaluation and deployment but says nothing about the verbalizer's description of that state. Thirdly, it situates NLAs within a longer history of verbal reports which are trained in conjunction with a process and then read as reports on that process (from chain of thought to human introspection) and draws on that history to give conditions under which such reports can be evidence.

An NLA description tells us which activation it refers to, but that, by itself, does not confirm if the claims about that activation are true. Because these descriptions are written so fluently, it is easy to miss this limitation, and to non-technical people, especially policymakers, this might seem like a good alignment and interpretability tool that would lead to transparency out of the box. The authors of the NLA know about this issue and use other methods to support their case studies. However, the types of behavior alignment that most need to be detected are often the ones where no other methods exist, so in these cases, there is no way to verify the description.

After 1977, psychology faced a similar issue with human introspection and then addressed it by limiting which reports could be used as data \citep{ericsson1980}. Similarly, we now suggest three restrictions for NLA output. First, the independence restriction means that some characterizations should be tested against the model’s behavior when it is changed, which the reconstructor makes cheap. Second, the reference requirement says that identifying and describing the content should be reported separately, so readers can see which part of the description the score supports. Third, the stakes criterion means that the rate of confabulation for claims that can be checked should be used as a lower bound on the prior probability of error for claims that cannot be checked. With these restrictions, NLAs will still serve as generators of hypotheses, and readers, philosophers, policy makers, and everyone using AI responsibly will be able to judge how well each hypothesis has been tested.

Kempelen's trick worked because the audience was looking inside and accepted the machinery as an adequate explanation for why the machine plays chess. Everything inside was real, the gears and the knobs, but it just had nothing to do with the real explanation. Unfortunately, even though they are an extremely useful tool, NLA descriptions invite the same misreading: they show, usually accurately, which state is active, and the audience takes them to say \textit{why} the model acted. NLAs are opening the cabinet and exposing the gears, but it is our task in explainable AI, alignment, philosophy, and cognitive science to check who is actually playing.

\bibliographystyle{plainnat}
\bibliography{references}

\end{document}